\documentclass[graybox]{svmult}

\usepackage{mathptmx}       
\usepackage{helvet}         
\usepackage{courier}        
\usepackage{type1cm}        
\usepackage{makeidx}         
\usepackage{graphicx}        
\usepackage{multicol}        
\usepackage[bottom]{footmisc}
\usepackage{verbatim}

\makeindex             

\begin{document}

\title*{Semantic Sentiment Analysis of Twitter Data}
\author{Preslav Nakov}
\institute{Preslav Nakov \at Qatar Computing Research Institute, HBKU, Doha, Qatar, \email{pnakov@qf.org.qa}
}
%
%
\maketitle


\section{Synonyms}

Microblog sentiment analysis; Twitter opinion mining

\section{Glossary}

	\textbf{Sentiment Analysis:} This is text analysis aiming to determine the attitude of a speaker or a writer with respect to some topic or the overall contextual polarity of a piece of text.

\section{Definition}

Sentiment analysis on Twitter is the use of natural language processing techniques to identify and categorize opinions expressed in a tweet, in order to determine the author's attitude toward a particular topic or in general. Typically, discrete labels such as \emph{positive}, \emph{negative}, \emph{neutral}, and \emph{objective} are used for this purpose, but it is also possible to use labels on an ordinal scale, or even continuous numerical values.

\section{Introduction}

Internet and the proliferation of smart mobile devices have changed the way information is created, shared, and spreads, e.g., microblogs such as Twitter, weblogs such as LiveJournal, social networks such as Facebook, and instant messengers such as Skype and WhatsApp are now commonly used to share thoughts and opinions about anything in the surrounding world.
This has resulted in the proliferation of social media content, thus creating new opportunities to study public opinion at a scale that was never possible before.

Naturally, this abundance of data has quickly attracted business and research interest from various fields including marketing, political science, and social studies, among many others, which are interested in questions like these: \emph{Do people like the new Apple Watch? What do they hate about iPhone6? Do Americans support ObamaCare? What do Europeans think of Pope's visit to Palestine? How do we recognize the emergence of health problems such as depression? Do Germans like how Angela Merkel is handling the refugee crisis in Europe? What do republican voters in USA like/hate about Donald Trump? How do Scottish feel about the Brexit?} 

Answering these questions requires studying the sentiment of opinions people express in social media, which has given rise to the fast growth of the field of sentiment analysis in social media, with Twitter being especially popular for research due to its scale, representativeness, variety of topics discussed, as well as ease of public access to its messages~\cite{java2007we,kwak2010twitter}.

Despite all these opportunities, the rise of social media has also
presented new challenges for natural language processing (NLP) applications,
which had largely relied on NLP tools tuned for formal text genres such as newswire,
and thus were not readily applicable to the informal language and style of social media.
That language proved to be quite challenging
with its use of creative spelling and punctuation, misspellings, slang, new words, URLs, and genre-specific terminology and abbreviations, e.g., RT for re-tweet and \#hashtags.
In addition to the genre difference, there is also a difference in length:
social media messages are generally short, often length-limited by design as in Twitter,
i.e., a sentence or a headline rather than a full document.
How to handle such challenges has only recently been the subject of thorough research
\cite{Barbosa10,Bifet11,Davidov:2010:SRS,Jansen09,Kouloumpis11,oconnor10,Pak10,Tumasjan10}.

\section{Key Points}

Sentiment analysis has a wide number of applications 
in areas such as market research, political and social sciences, 
and for studying public opinion in general,
and Twitter is one of the most commonly-used platforms for this.
This is due to its streaming nature, which allows for real-time analysis,
to its social aspect, which encourages people to share opinions, and
to the short size of the tweets, which simplifies linguistic analysis.

\noindent There are several formulations of the task of Sentiment Analysis on Twitter
that look at different sizes of the target (e.g., at the level of words vs. phrases vs. tweets vs. sets of tweets),
at different types of semantic targets (e.g., aspect vs. topic vs. overall tweet),
at the explicitness of the target (e.g., sentiment vs. stance detection),
at the scale of the expected label (2-point vs. 3-point vs. ordinal), etc.
All these are explored at SemEval, the International Workshop on Semantic Evaluation, which has created a number of benchmark datasets and has enabled direct comparison between different systems and approaches, both as part of the competition and beyond.

Traditionally, the task has been addressed using supervised and semi-supervised methods, as well as using distant supervision,
with the most important resource being sentiment polarity lexicons,
and with feature-rich approaches as the dominant research direction for years.
With the recent rise of deep learning, which in many cases eliminates the need for any explicit feature modeling, the importance of both lexicons and features diminishes, while at the same time attention is shifting towards learning from large unlabeled data, which is needed to train the high number of parameters of such complex models. 
Finally, as methods for sentiment analysis mature, more attention is also being paid to linguistic structure and to multi-linguality and cross-linguality.

\section{Historical Background}

Sentiment analysis emerged as a popular research direction in the early 2000s.
Initially, it was regarded as standard document classification into topics such as business, sport, and politics \cite{Pang:2002:TUS}.
However, researchers soon realized that it was quite different from standard document classification \cite{Sebastiani:2002:MLA:survey},
and that it crucially needed external knowledge in the form of sentiment polarity lexicons.

Around the same time, other researchers realized the importance of external sentiment lexicons, e.g., Turney~\cite{turney2002thumbs} proposed an unsupervised approach to learn the sentiment orientation of words/phrases: positive vs. negative. Later work studied the linguistic aspects of expressing opinions, evaluations, and speculations \cite{Wiebe:2004:LSL}, the role of context in determining the sentiment orientation \cite{Wilson05}, of deeper linguistic processing such as negation handling \cite{PangL08}, of finer-grained sentiment distinctions \cite{Pang2005}, 
of positional information \cite{raychev-nakov:2009:RANLP09}, 
etc. Moreover, it was recognized that in many cases, it is crucial to know not just the polarity of the sentiment but also the topic toward which this sentiment is expressed \cite{Stoyanov:2008:TIF}.

Until the rise of social media, research on opinion mining and sentiment analysis
had focused primarily on learning about the language of sentiment in general,
meaning that it was either genre-agnostic \cite{swn}
or focused on newswire texts \cite{Wiebe05} and customer reviews (e.g., from web forums),
most notably about movies \cite{Pang:2002:TUS} and restaurants \cite{Semeval2014task4}
but also about hotels, digital cameras, cell phones, MP3 and DVD players \cite{Hu04},
laptops \cite{Semeval2014task4},
etc.
This has given rise to several resources, mostly word and phrase polarity lexicons,
which have proven to be very valuable for their respective domains and types of texts,
but less useful for short social media messages.

\noindent Later, with the emergence of social media, sentiment analysis in Twitter became a hot research topic.
Unfortunately, research in that direction was hindered by the unavailability of suitable datasets and lexicons for system training, development, and testing.
While some Twitter-specific resources were developed,
initially they were either small and proprietary, such as the i-sieve corpus \cite{Kouloumpis11},
were created only for Spanish like the TASS corpus \cite{TASS},
or relied on noisy labels obtained automatically, e.g., 
based on emoticons and hashtags 
\cite{mohammad:2012:STARSEM-SEMEVAL,MohammadKZ2013,Pang:2002:TUS}.

This situation changed with the shared task on \emph{Sentiment Analysis on Twitter}, which was organized at SemEval, the International Workshop on Semantic Evaluation, a semantic evaluation forum previously known as SensEval. The task ran in 2013, 2014, 2015, and 2016, attracting over 40 participating teams in all four editions. While the focus was on general tweets, the task also featured out-of-domain testing on SMS messages, LiveJournal messages, as well as on sarcastic tweets. 

SemEval-2013 Task 2 \cite{Semeval2013} and SemEval-2014 Task 9 \cite{rosenthal-EtAl:2014:SemEval} focused on expression-level and message-level polarity.
SemEval-2015 Task 10 \cite{Nakov:2016rm,Rosenthal-EtAl:2015:SemEval} featured topic-based message polarity classification on detecting trends toward a topic and on determining the out-of-context (a priori) strength of association of Twitter terms with positive sentiment.
SemEval-2016 Task 4 \cite{nakov-EtAl:2016:SemEval} introduced a 5-point scale, which is used for human review ratings on popular websites such as Amazon, TripAdvisor, Yelp, etc.; from a research perspective, this meant moving from classification to \emph{ordinal regression}.
Moreover, it focused on \emph{quantification}, i.e., determining what proportion of a set of tweets on a given topic are positive/negative about it. It also featured a 5-point scale \emph{ordinal quantification} subtask \cite{Gao:2015ly}.

Other related 
tasks have explored 
aspect-based sentiment analysis \cite{pontiki-EtAl:2015:SemEval,SemEval:2016:task5,Semeval2014task4},
sentiment analysis of figurative language on Twitter \cite{ghosh-EtAl:2015:SemEval},
implicit event polarity \cite{russo-caselli-strapparava:2015:SemEval},
stance in tweets \cite{SemEval:2016:task6},
out-of-context sentiment intensity of phrases \cite{SemEval:2016:task7},
and emotion detection \cite{SemEval2007}.
Some of these tasks featured languages other than English.

\section{Sentiment Analysis on Twitter: A SemEval Perspective}

\subsection{Variants of the Task at SemEval}

\textbf{Tweet-level sentiment.} The simplest and also the most popular task of sentiment analysis on Twitter is to determine the overall sentiment expressed by the author of a tweet \cite{nakov-EtAl:2016:SemEval,Nakov:2016rm,Semeval2013,Rosenthal-EtAl:2015:SemEval,rosenthal-EtAl:2014:SemEval}.
Typically, this means choosing one of the following three classes to describe the sentiment: \textsc{Positive}, \textsc{Negative}, and \textsc{Neutral}. Here are some examples:

\begin{enumerate}
	\item[(1)]  \textsc{Positive}: \textit{@nokia lumia620 cute and small and pocket-size, and available in the brigh
	 test colours of day! \#lumiacaption}
	\item[(2)]  \textsc{Negative}: \textit{I hate tweeting on my iPhone 5 it's so small :(}
	\item[(3)] \textsc{Neutral}: \textit{If you work as a security in a samsung store...Does that make you guardian of the galaxy??}
\end{enumerate}

\textbf{Sentiment polarity lexicons.} Naturally, the overall sentiment in a tweet can be determined based on the sentiment-bearing words and phrases it contains as well as based on emoticons such as \texttt{;)} and\texttt{:(}. For this purpose, researchers have been using lexicons of sentiment-bearing words.
For example, \emph{cute} is a positive word, while \emph{hate} is a negative one, and the occurrence of these words in (1) and (2) can help determine the overall polarity of the respective tweet. We will discuss these lexicons in more detail below.

\textbf{Prior sentiment polarity of multi-word phrases.} Unfortunately, many sentiment-bearing words are not universally good or universally bad. For example, the polarity of an adjective could depend on the noun it modifies, e.g., \emph{hot coffee} and \emph{unpredictable story} express positive sentiment, while \emph{hot beer} and \emph{unpredictable steering} are negative. Thus, determining the out-of-context (a priori) strength of association of Twitter terms, especially multi-word terms, with positive/negative sentiment is an active research direction \cite{Nakov:2016rm,Rosenthal-EtAl:2015:SemEval}. 

\textbf{Phrase-level polarity in context.} Even when the target noun is the same, the polarity of the modifying adjective could be different in different tweets, e.g., \emph{small} is positive in (1) but negative in (2), even though they both refer to a phone. Thus, there has been research in determining the sentiment polarity of a term in the context of a tweet \cite{Semeval2013,Rosenthal-EtAl:2015:SemEval,rosenthal-EtAl:2014:SemEval}.

\textbf{Sarcasm.} Going back to tweet-level sentiment analysis, we should mention sarcastic tweets, which are particularly challenging as the sentiment they express is often the opposite of what the words they contain suggest \cite{Davidov:2010:SRS,Rosenthal-EtAl:2015:SemEval,rosenthal-EtAl:2014:SemEval}. For example, (4) and (5) express a negative sentiment even though they contain positive words and phrases such as \emph{thanks}, \emph{love}, and \emph{boosts my morale}.

\begin{enumerate}
	\item[(4)]  \textsc{Negative}: \textit{Thanks manager for putting me on the schedule for Sunday}
	\item[(5)]  \textsc{Negative}: \textit{I just love missing my train every single day. Really boosts my morale.}
\end{enumerate}

\textbf{Sentiment toward a topic.} Even though tweets are short, as they are limited to 140 characters by design (even though this was relaxed a bit as of September 19, 2016, and now media attachments such as images, videos, polls, etc., and quoted tweets no longer reduce the character count),
they are still long enough to allow the tweet's author to mention several topics and to express potentially different sentiment toward each of them. 
A topic can be anything that people express opinions about, e.g., a product (e.g., \emph{iPhone6}), a political candidate (e.g., \emph{Donald Trump}), a policy (e.g.,~\emph{Obamacare}), an event (e.g.,~\emph{Brexit}), etc.
For example, in (6) the author is positive about Donald Trump but negative about Hillary Clinton. A political analyzer would not be interested so much in the overall sentiment expressed in the tweet (even though one could argue that here it is positive overall), but rather in the sentiment with respect to a topic of his/her interest of study.

\begin{enumerate}
	\item[(6)] \textit{As a democrat I couldnt ethically support Hillary no matter who was running against her. Just so glad that its Trump, just love the guy!}\\
	 (topic: \emph{Hillary} $\rightarrow$ \textsc{Negative})\\
	 (topic: \emph{Trump} $\rightarrow$ \textsc{Positive})
\end{enumerate}


\textbf{Aspect-based sentiment analysis.} Looking again at (1) and (2), we can say that the sentiment is not about the phone (\emph{lumia620} and \emph{iPhone 5}, respectively), but rather about some specific aspect thereof, namely, \textsc{size}. Similarly, in (7) instead of sentiment toward the topic \emph{lasagna}, we can see sentiment toward two aspects thereof: \textsc{quality} (\textsc{Positive} sentiment) and \textsc{quantity} (\textsc{Negative} sentiment). Aspect-based sentiment analysis is an active research area \cite{pontiki-EtAl:2015:SemEval,SemEval:2016:task5,Semeval2014task4}.

\begin{enumerate}
	\item[(7)] \textit{The lasagna is delicious but do not come here on an empty stomach.}
\end{enumerate}

\textbf{Stance detection.} A task related to, but arguably different in some respect from sentiment analysis, is that of \emph{stance detection}. The goal here is to determine whether the author of a piece of text is in favor of, against, or neutral toward a proposition or a target \cite{SemEval:2016:task6}.
For example, in (8) the author has a negative stance toward the proposition \emph{w​omen have the right to abortion}, even though the target is not mentioned at all. Similarly, in (9§) the author expresses a negative sentiment toward \emph{Mitt Romney}, from which one can imply that s/he has a positive stance toward the target \emph{​Barack Obama}.

\begin{enumerate}
	\item[(8)] \textit{A​ foetus has rights too! Make your voice heard.}\\
	(Target: \emph{w​omen have the right to abortion} $\rightarrow$ \textsc{Against})
\end{enumerate}

\begin{enumerate}
	\item[(9)] \textit{A​ll Mitt Romney cares about is making money for the rich.}\\
	(Target: \emph{​Barack Obama} $\rightarrow$ \textsc{InFavor})
\end{enumerate}

\textbf{Ordinal regression.} The above tasks were offered in different granularities, e.g., 2-way (\textsc{Positive}, \textsc{Negative}), 3-way (\textsc{Positive}, \textsc{Neutral},
\textsc{Negative}), 4-way (\textsc{Positive}, \textsc{Neutral}, \textsc{Negative}, \textsc{Objective}), 5-way (\textsc{HighlyPositive}, \textsc{Positive}, \textsc{Neutral}, \textsc{Negative}, \textsc{HighlyNegative}), and sometimes even 11-way \cite{ghosh-EtAl:2015:SemEval}. It is important to note that the 5-way and the 11-way scales are ordinal, i.e., the classes can be associated with numbers, e.g., $-$2, $-$1, 0, 1, and 2 for the 5-point scale. This changes the machine learning task as not all mistakes are equal anymore \cite{Pang2005}. For example, misclassifying a \textsc{HighlyNegative} example as \textsc{HighlyPositive} is a bigger mistake than misclassifying it as \textsc{Negative} or as \textsc{Neutral}. From a machine learning perspective, this means moving from \emph{classification} to \emph{ordinal regression}. This also requires different evaluation measures \cite{nakov-EtAl:2016:SemEval}.

\textbf{Quantification.} 
Practical applications are hardly ever interested in the sentiment expressed in a \emph{specific tweet}. Rather, they look at estimating the \textit{prevalence} of positive
  and negative tweets about a given topic in a set of tweets from some time interval. 
Most (if not all) tweet sentiment classification studies conducted within political science
  \cite{Borge-Holthoefer:2015dz,Kaya:2013ca,marchettibowick-chambers:2012:EACL2012},
  economics \cite{Bollen:2011bf,oconnor10}, social science
  \cite{Dodds:2011uq}, and market research
  \cite{Burton:2011sh,Qureshi:2013fb} use Twitter with an interest in
  aggregate data and \emph{not} in individual classifications.
  Thus, some tasks, such as SemEval-2016 Task 4 \cite{nakov-EtAl:2016:SemEval}, replace classification with class prevalence estimation, which is also known as \textit{quantification} in data mining and related fields. Note that quantification is not a mere byproduct of classification, since a good classifier is not necessarily a good quantifier, and vice versa \cite{Forman:2008kx}.
  Finally, in case of multiple labels on an ordinal scale, we have yet another machine learning problem: \emph{ordinal quantification}. Both versions of quantification require specific evaluation measures and machine learning algorithms.

\subsection{Features and Learning}

\textbf{Pre-processing.} Tweets are subject to standard preprocessing steps for text such as tokenization, stemming, lemmatization, stop-word removal, and part-of-speech tagging. Moreover, due to their noisy nature, they are also processed using some Twitter-specific techniques such as substitution/removal of URLs, of user mentions, of hashtags, and of emoticons, spelling correction, elongation normalization, abbreviation lookup, punctuation removal, detection of amplifiers and diminishers, negation scope detection, etc. 
For this, one typically uses Twitter-specific NLP tools such as part-of-speech and named entity taggers, syntactic parsers, etc. \cite{gimpel2011,kong-EtAl:2014:EMNLP2014,ner}.

\textbf{Negation handling.} Special handling is also done for negation.
The most popular approach to negation handling is to transform any word that appeared in a negation context by adding a suffix \textit{\_NEG} to it, e.g., \textit{good} would become \textit{good\_NEG} \cite{Das2001,Pang:2002:TUS}.
A negated context is typically defined as a text span between a negation word, e.g., {\it no, not, shouldn't}, and a punctuation mark or the end of the message. Alternatively, one could flip the polarity of sentiment words, e.g., the positive word \textit{good} would become negative when negated. It has also been argued \cite{Zhu2014} that negation affects different words differently, and thus it was also proposed to build and use special sentiment polarity lexicons for words in negation contexts \cite{Kiritchenko2014}.

\textbf{Features.} Traditionally, systems for Sentiment Analysis on Twitter have relied on handcrafted features derived from word-level (e.g., \emph{great}, \emph{freshly roasted coffee}, \emph{becoming president}) and character-level $n$-grams (e.g., \emph{bec}, \emph{beco}, \emph{comin}, \emph{oming}), stems (e.g., \emph{becom}), lemmata (e.g., \emph{become}, \emph{roast}), punctuation (e.g., exclamation and question marks), part-of-speech tags (e.g.,~adjectives, adverbs, verbs, nouns), word clusters (e.g., \emph{probably}, \emph{probly}, and \emph{maybe} could be collapsed to the same word cluster), and Twitter-specific encodings such as emoticons (e.g., \emph{;)}, \emph{:D}), hashtags (\emph{\#Brexit}), user tags (e.g., \verb!@!\emph{allenai\_org}), abbreviations (e.g., \emph{RT}, \emph{BTW}, \emph{F2F}, \emph{OMG}), elongated words (e.g., \emph{soooo}, \emph{yaayyy}), use of capitalization (e.g., proportion of ALL CAPS words), URLs, etc.
Finally, the most important features are those based on the presence of words and phrases in sentiment polarity lexicons with positive/negative scores; examples of such features include number of positive terms, number of negative terms, ratio of the number of positive terms to the number of positive+negative terms, ratio of the number of negative terms to the number of positive+negative terms, sum of all positive scores, sum of all negative scores, sum of all scores, etc.

\textbf{Supervised learning.} Traditionally, the above features were fed into classifiers such as Maximum Entropy (MaxEnt) and Support Vector Machines (SVM) with various kernels. However, observation over the SemEval Twitter sentiment task in recent years shows growing interest in, and by now clear dominance of methods based on deep learning. In particular, the best-performing systems at SemEval-2015 and SemEval-2016 used deep convolutional networks \cite{SemEval:2016:task4:SwissCheese,Severyn:2015:TSA:2766462.2767830}. Conversely, kernel machines seem to be less frequently used than in the past, and the use of learning methods other than the ones mentioned above is at this point scarce. All these models are examples of supervised learning as they need labeled training data.

\textbf{Semi-supervised learning.} We should note two things about the use of deep neural networks. First they can often do quite well without the need for explicit feature modeling, as they can learn the relevant features in their hidden layers starting from the raw text. Second, they have too many parameters, and thus they require a lot of training data, orders of magnitude more than it is realistic to have manually annotated. A popular way to solve this latter problem is to use self training, a form of semi-supervised learning, where first a system is trained on the available training data only, then this system is applied to make predictions on a large unannotated set of tweets, and finally it is trained for a few more iterations on its own predictions. This works because parts of the network, e.g., with convolution or with LSTMs \cite{dossantos-gatti:2014:Coling,Severyn:2015:TSA:2766462.2767830,wang-EtAl:2015:ACL-IJCNLP2}, need to learn something like a language model, i.e., which word is likely to follow which one. Training these parts needs no labels. While these parts can be also pre-trained, it is easier, and often better, to use self training.

\textbf{Distantly-supervised learning.} Another way to make use of large unannotated datasets is to rely on \emph{distant supervision} \cite{marchettibowick-chambers:2012:EACL2012}. For example, one can annotate tweets for sentiment polarity based on whether they contain a positive or a negative emoticon. This results in noisy labels, which can be used to train a system \cite{Severyn:2015:TSA:2766462.2767830}, to induce sentiment-specific word embeddings \cite{tang-EtAl:2014:P14-1}, sentiment-polarity lexicons \cite{MohammadKZ2013}, etc.

\textbf{Unsupervised learning.} Fully unsupervised learning is not a popular method for addressing sentiment analysis tasks. Yet, some features used in sentiment analysis have been learned in an unsupervised way, e.g., Brown clusters to generalize over words \cite{Owoputi12part-of-speechtagging}. Similarly, word embeddings are typically trained from raw tweets that have no annotation for sentiment (even though there is also work on sentiment-specific word embeddings \cite{tang-EtAl:2014:P14-1}, which uses distant supervision).

\section{Sentiment Polarity Lexicons}

Despite the wide variety of knowledge sources explored so far in the literature, sentiment polarity lexicons remain the most commonly used resource for the task of sentiment analysis.


Until recently, such sentiment polarity lexicons were manually crafted and were thus of small to moderate size, e.g.,
LIWC \cite{pennebaker01} has 2,300 words,
the General Inquirer \cite{inquirer1966computer} contains 4,206 words,
Bing Liu's lexicon \cite{Hu04} includes 6,786 words,
and MPQA \cite{Wilson05} has about 8,000 words.


Early efforts toward building sentiment polarity lexicons automatically yielded lexicons of moderate sizes such as the SentiWordNet \cite{swn,esuli:lrec2006}.
However, recent results have shown that automatically extracted large-scale lexicons (e.g., up to a million words and phrases) offer important performance advantages, as confirmed at shared tasks on Sentiment Analysis on Twitter at SemEval 2013-2016 \cite{nakov-EtAl:2016:SemEval,Semeval2013,Rosenthal-EtAl:2015:SemEval,rosenthal-EtAl:2014:SemEval}.
Using such large-scale lexicons was crucial for the performance of the top-ranked systems. Similar observations were made in the related Aspect-Based Sentiment Analysis task at SemEval 2014 \cite{Semeval2014task4}. 
In both tasks, the winning systems benefitted from building and using massive sentiment polarity lexicons 
\cite{MohammadKZ2013,Zhu_SemEval2014}.

\noindent The two most popular large-scale lexicons were the Hashtag Sentiment Lexicon and the Sentiment140 lexicon, which were developed by the team of NRC Canada for their participation in the SemEval-2013 shared task on sentiment analysis on Twitter.
Similar automatically induced lexicons proved useful for other SemEval tasks, e.g., for SemEval-2016 Task 3 on Community Question Answering \cite{SemEval2016:task3:PMI-cool,nakov-EtAl:2016:SemEval}.
 
The importance of building sentiment polarity lexicons has resulted in a special subtask \cite{Rosenthal-EtAl:2015:SemEval} at SemEval-2015 (part of Task 4) and an entire task \cite{SemEval:2016:task7} at SemEval-2016 (namely, Task 7), on predicting the out-of-context sentiment intensity of words and phrases.
Yet, we should note though that the utility of using sentiment polarity lexicons for sentiment analysis probably needs to be revisited, as the best system at SemEval-2016 Task 4 could win without using any lexicons \cite{SemEval:2016:task4:SwissCheese}; it relied on semi-supervised learning using a deep neural network.



Various approaches have been proposed in the literature
for bootstrapping sentiment polarity lexicons starting from a small set of seeds:
positive and negative terms (words and phrases).
The dominant approach is that of Turney \cite{turney2002thumbs}, who uses pointwise mutual information and bootstrapping to build a large lexicon and to estimate the semantic orientation of each word in that lexicon.
He starts with a small set of seed positive (e.g., \emph{excellent}) and negative words (e.g., \emph{bad}), and then uses these words to induce sentiment polarity orientation for new words in a large unannotated set of texts (in his case, product reviews). The idea is that words that co-occur in the same text with positive seed words are likely to be positive, while those that tend to co-occur with negative words are likely to be negative. To quantify this intuition, Turney defines the notion of sentiment orientation (SO) for a term $w$ as follows:

\begin{center}
\textbf{$SO(w) = pmi(w,pos) - pmi(w,neg)$}
\end{center}

\noindent where PMI is the pointwise mutual information, $pos$ and $neg$ are placeholders standing for any of the seed positive and negative terms, respectively, and $w$ is a target word/phrase from the large unannotated set of texts (here tweets).

A positive/negative value for $SO(w)$ indicates positive/negative polarity for the word $w$, and its magnitude shows the corresponding sentiment strength. In turn, $pmi(w,pos)=\frac{P(w,pos)}{P(w)P(pos)}$, where $P(w,pos)$ is the probability to see $w$ with any of the seed positive words in the same tweet,
$P(w)$ is the probability to see $w$ in any tweet, and $P(pos)$ is the probability to see any of the seed positive words in a tweet;  $pmi(w,neg)$ is defined similarly.

The pointwise mutual information is a notion from information theory: given two random variables $A$ and $B$, the mutual information of $A$ and $B$ is the ``amount of information'' (in units such as bits) obtained about the random variable $A$, through the random variable $B$ \cite{Church:1990:WAN:89086.89095}. 

Let $a$ and $b$ be two values from the sample space of $A$ and $B$, respectively. The \emph{pointwise} mutual information between $a$ and $b$ is defined as follows:
\begin{equation}
pmi(a;b) = \log{\frac{P(A = a, B = b)}{P(A=a)\cdot P(B=b)}} = \log{\frac{P(A=a|B=b)}{P(A=a)}} 
\end{equation}

$pmi(a;b)$ takes values between $-\infty$, which happens when $P(A=a,B=b)$ = 0, and 
$\min\left\{ -\log P(A=a), -\log P(B=b) \right\}$ if $P(A=a|B=b) = P(B=b|A=a) = 1$.



In his experiments, Turney \cite{turney2002thumbs} used five positive and five negative words as seeds.
His PMI-based approach further served as the basis for the creation of the two above-mentioned large-scale automatic lexicons for sentiment analysis in Twitter for English,
initially developed by NRC for their participation in SemEval-2013 \cite{MohammadKZ2013}.
The \emph{Hashtag Sentiment Lexicon} uses as seeds hashtags containing 32 positive and 36 negative words, e.g.,  \texttt{\#happy} and \texttt{\#sad}.
Similarly, the \emph{Sentiment140} lexicon uses smileys as seed indicators for positive and negative sentiment, e.g.,  \texttt{:)}, \texttt{:-)}, and \texttt{:))} as positive seeds, and \texttt{:(} and \texttt{:-(} as negative ones.

An alternative approach to lexicon induction has been proposed \cite{severyn2015automatic}, which, instead of using PMI, assigns positive/negative labels to the unlabeled tweets (based on the seeds), and then trains an SVM classifier on them, using word $n$-grams as features. These $n$-grams are then used as lexicon entries (words and phrases) with the learned classifier weights as polarity scores.
Finally, it has been shown that sizable further performance gains can be obtained by starting with mid-sized seeds, i.e., hundreds of words and phrases
\cite{jovanoski-pachovski-nakov:2016:COLING2016}.

\section{Key Applications}

Sentiment analysis on Twitter has applications in a number of areas, including 
political science \cite{Borge-Holthoefer:2015dz,Kaya:2013ca,marchettibowick-chambers:2012:EACL2012},
economics \cite{Bollen:2011bf,oconnor10},
social science \cite{Dodds:2011uq}, 
and market research \cite{Burton:2011sh,Qureshi:2013fb}.
It is used to study company reputation online \cite{Qureshi:2013fb}, to measure customer satisfaction, to identify detractors and promoters, to forecast market growth \cite{Bollen:2011bf}, to predict the future income from newly-released movies, to forecast the outcome of upcoming elections \cite{marchettibowick-chambers:2012:EACL2012,oconnor10}, to study political polarization \cite{Borge-Holthoefer:2015dz,Tumasjan10}, etc.

\section{Future Directions}

We expect the quest for more interesting formulations of the general sentiment analysis task to continue. We see competitions such as those at SemEval as the engine of this innovation, as they not only perform head-to-head comparisons, but also create databases and tools that enable follow-up research for many years afterward.

In terms of methods, we believe that deep learning \cite{dossantos-gatti:2014:Coling,Severyn:2015:TSA:2766462.2767830,wang-EtAl:2015:ACL-IJCNLP2}, together with semi-supervised and distantly-supervised methods \cite{Davidov:2010:ESL:1944566.1944594,tang-EtAl:2014:P14-1}, will be the main focus of future research. We also expect more attention to be paid to linguistic structure and sentiment compositionality \cite{Socher:2012:SCT:2390948.2391084,socher-EtAl:2013:EMNLP}.
Moreover, we forecast more interest for languages other than English, and for cross-lingual methods \cite{Kaya:2013ca,mihalcea-banea-wiebe:2007:ACLMain,Wan:2009:CCS:1687878.1687913}, which will allow leveraging on the rich resources that are already available for English.
Last, but not least, the increase in opinion spam on Twitter will make it important to study astroturfing \cite{Ratkiewicz:2011:TMS:1963192.1963301} and troll detection \cite{mihaylov-georgiev-nakov:2015:CoNLL,mihaylov-EtAl:2015:RANLP2015,mihaylov-nakov:2016:P16-2}.

\section{Cross-References}

Microblog Sentiment Analysis	100590

\noindent Multi-classifier System for Sentiment Analysis and Opinion Mining	351

\noindent Sentiment Analysis in Social Media	120

\noindent Sentiment Analysis of Microblogging Data	110168

\noindent Sentiment Analysis of Reviews	110169

\noindent Sentiment Analysis, Basics of	110159

\noindent Sentiment Quantification of User-Generated Content	110170

\noindent Social Media Analysis for Monitoring Political Sentiment	110172

\noindent Twitter Microblog Sentiment Analysis	265

\noindent User Sentiment and Opinion Analysis	192



\bibliographystyle{spmpsci}
\bibliography{bib}

\section{Recommended Reading}

For general research on sentiment analysis, we recommend the following surveys: \cite{Liu12} and \cite{PangL08}. For sentiment analysis on Twitter, we recommend the overview article on \emph{Sentiment Analysis on Twitter} about the SemEval task \cite{Nakov:2016rm} as well as the task description papers for different editions of the task \cite{nakov-EtAl:2016:SemEval,Semeval2013,Rosenthal-EtAl:2015:SemEval,rosenthal-EtAl:2014:SemEval}.

\end{document}